\documentclass[conference]{IEEEtran}
\IEEEoverridecommandlockouts
\usepackage{cite}
\usepackage{amsmath,amssymb,amsfonts}
\usepackage{graphicx}
 \usepackage{float}
\usepackage[caption=false,font=footnotesize]{subfig}
\usepackage{textcomp}
\usepackage{xcolor}
\usepackage{booktabs}
\usepackage{multirow}
\usepackage{url}
\usepackage{hyperref}
\usepackage{tikz}
\usepackage{pgf}
\usetikzlibrary{arrows.meta,positioning,fit,backgrounds,shapes.geometric}
\def\BibTeX{{\rm B\kern-.05em{\sc i\kern-.025em b}\kern-.08em
    T\kern-.1667em\lower.7ex\hbox{E}\kern-.125emX}}
\begin{document}
\title{How to Leverage Synthetic Speech for LLM-Based ASR Systems?}
\author{
\IEEEauthorblockN{
Yanis Labrak$^{1}$,
Dairazalia Sanchez-Cortes$^{1}$,
Sergio Burdisso$^{1}$, 
Séverin Baroudi$^{2}$, 
Shashi Kumar$^{1, 3}$, \\
Esaú Villatoro-Tello$^{1}$,
Srikanth Madikeri$^{4}$,
Manjunath K E$^{5}$,
Oldřich Plchot$^{6}$, \\
Kadri Hacio\u{g}lu$^{5}$,
Petr Motlicek$^{1, 6}$,
Andreas Stolcke$^{5}$
}
\vspace{3mm}
\IEEEauthorblockA{
$^{1}$\textit{Idiap Research Institute, Martigny, Switzerland} \\
$^{2}$\textit{Universite de Toulon, Aix Marseille Univ, LIS, CNRS, Toulon, France} \\
$^{3}$\textit{EPFL, Lausanne, Switzerland};
$^{4}$\textit{University of Zurich, Zurich, Switzerland} \\
$^{5}$\textit{Uniphore, U.S.A. \& India};
$^{6}$\textit{Brno University of Technology, Brno, Czech Republic}
}
}
\maketitle
\begin{abstract}
In regulated domains such as banking and healthcare, where privacy makes real speech costly to collect, synthetic speech from text-to-speech is an appealing alternative for training automatic speech recognition (ASR). Yet a distributional gap limits how far synthetic data can replace real recordings. While prior work treats this gap as a black box to engineer around, we probe its origin in a SLAM-ASR architecture. We localise the signal separating real from synthetic speech to early-to-middle layers of the LLM backbone, where temporal and prosodic perturbations disrupt it most. Representation-level separability helps but does not directly predict downstream ASR gains. Convolving synthetic audio with room impulse responses (RIR) narrows the gap not by making speech more natural, but by reproducing the acoustic irregularities of real recordings. Combining a layer-selection module with RIR augmentation matches an all-real baseline using only 25\% of real speech (13.6h) and surpasses it at all higher proportions.

\end{abstract}

\section{Introduction}

Speech data in regulated domains such as banking and healthcare is among the hardest to collect and retain at scale. Customer--agent telephone calls contain personally identifiable and financial information, and data-protection regimes such as the GDPR \cite{eu2024aiact} constrain how such recordings may be stored, shared, and reused for model training, since voice recordings potentially qualifying as biometric data. Synthetic speech from modern text-to-speech (TTS) offers an appealing way around these constraints, utterances can be generated on demand without exposing real customers, sidestepping the collection and retention of sensitive audio. This appealing promise, could help to cut the cost and the privacy burden of annotating real speech for ASR. However, it only holds if synthetic speech can actually substitute for real recordings, which in turn requires closing the distributional gap that still separates the two. Most, modern LLM-based ASR models still separate real from synthetic signals with high accuracy \cite{guo2026explicitacousticevidenceperception}, limiting how far synthetic audio can replace real recordings in low-resource settings. Rather than treating this gap as a black box, we examine where it arises inside the model and use that to guide practical training recipes. We organise the study around four research questions:

\begin{enumerate}
    \item \textbf{Where is real/synthetic discrimination encoded?} Which layers of the LLM backbone most strongly separate the two classes and can be leveraged using layer-wise weighted pooling (Figure \ref{fig:lwp_arch})? At the same time, which  signal-level perturbations disrupt it most?
    \item \textbf{Do interpretability-guided filters improve ASR?} Does disrupting the real/synthetic boundary at the representation level translate to downstream WER gains?
    \item \textbf{How much real data can synthetic replace?} What real/synthetic mix preserves or improves over an all-real baseline?
    \item \textbf{How much synthetic data helps on top of full real data?} Does raw or RIR-augmented synthetic audio help when added to a augment real corpus?
\end{enumerate}

Taken together, our findings provide one of the first answer to these questions. Probing the LoRA-adapted LLM backbone with overlap metrics (Section \ref{subsection:overlap_metrics}), we localise where the synthetic/real gap is concentrated and which signal-level perturbations affect it most (Section~\ref{sec:ablation}). We then show that this representation-level separability does not necessarily predict downstream gains while guiding us (Section \ref{tab:rq5}), and that room impulse response (RIR) convolution narrows the gap not by improving perceptual quality but by reproducing the acoustic irregularities of real recordings (Section~\ref{sec:downstream}). Guided by these findings, a per-token layer-wise weighted pooling over the decoder layers (Figure~\ref{fig:lwp_arch}), combined with RIR augmentation, matches a fully real-data baseline using only 25\% of the real speech and surpasses it at all higher proportions (Section~\ref{sec:lwp}).

\section{Background and Related Work}

\subsection{Speech LLMs and LLM-based ASR}

Coupling a speech encoder to a pre-trained large language model (LLM) through a trainable projector has become a dominant recipe for general purpose audio understanding, we refer to this family as SpeechLLM. Systems such as SALMONN~\cite{tang2024salmonn}, Qwen-Audio~\cite{chu2023qwenaudioadvancinguniversalaudio}, and AudioPaLM~\cite{rubenstein2023audiopalmlargelanguagemodel} connect self-supervised or supervised encoders to a frozen or slightly adapted text LLM, reaching strong ASR performance alongside broader range of audio tasks. SLAM-ASR~\cite{ma2024slam}, shows that a single trainable projector between a frozen WavLM~\cite{chen2022wavlm} encoder and a frozen LLM suffices for competitive performances, in contrast to fully supervised encoder-decoder models such as Whisper~\cite{radford2023whisper}. Work on these models has concentrated on linguistic performance, but, understanding how the LLM backbone encodes non-linguistic properties, remains largely unexamined \cite{lu2026auditoryknowledge}.

\subsection{Training ASR with Synthetic Speech and Its Limits}

TTS generated speech is widely used to augment low-resource ASR~\cite{rossenbach2020ttsaugmentation, rosenberg2019synth, wang21r_interspeech}, and modern multi-speaker TTS has made this a practical foundation to their works~\cite{yang2024versatiletts, perrin2025optsynth}. The benefits, however, are uneven, and two obstacles remains not examined properly. First, performances are often in those papers dependent on the amount of data generated~\cite{yang2024versatiletts} and not the diversity of its voices or the acoustic conditions. Second, a systematic acoustic mismatch separates synthetic from real speech, open-source TTS outputs are cleaner and more uniform than real recordings~\cite{srinivasavaradhan2025ttsstate}, so a model trained on it learns cues specific to the model and transfers poorly to real audio~\cite{rosenberg2019synth, su2024taskarith}.
Recent methods reduce this mismatch in several ways: by merging the weights of models trained separately on real and synthetic data~\cite{su2024taskarith}, by filtering and optimising the generation pipeline~\cite{perrin2025optsynth}, or by letting ASR and TTS models refine one another in a closed loop~\cite{chou2025selfrefining}.
All of these act on the data, the TTS model, or the final weights of the ASR model, all considering the model's training procedure as a black box. None of this previous work, raised questions about the where inside that model does the synthetic/real distinction is actually represented.

\subsection{Interpretability of Speech Models}

Modern neural vocoders and diffusion models have pushed TTS perceptual quality toward human level, yet residual cues in pitch, prosody, and spectral envelope persist~\cite{tan2022naturalspeech, mishra2026explainable}. These cues constitute a distributional gap between synthetic and real speech that perceptual fidelity alone does not close. The anti-spoofing community studies precisely this gap, the ASVspoof initiative~\cite{wang2020asvspoof} and subsequent detectors~\cite{dvirniak2026robustspeechdeepfakedetection} show that synthetic speech is reliably detectable because it carries systematic, learnable signatures. The detectability that anti-spoofing exploits and the mismatch that limits augmentation are thus two views of the same discrepancy.
What neither line of work asks is where, inside a speech model, that gap is encoded. 
The closest evidence comes from interpretability work, which has surfaced cues about how speech models represent information yet has centered on the encoder\cite{baroudi24_interspeech, zaiem23b_interspeech, 11434629}, layer-wise probing reveals a progression from acoustic to semantic features across encoder layers~\cite{pasad2021layerwise}, and aggregating those layers through a learnable weighted sum, as popularised by SUPERB~\cite{yang2021superb}, is now standard practice. Attention has only recently turned to the LLM backbone analysis that consumes these features for understanding tasks, focusing in particular on the drop in performance between text and speech inputs~\cite{hsu2026anatomymodalitygapdissecting}, rather than to acoustic distinctions within speech itself. To our knowledge, no prior work examines how a speech LLM internally represents the synthetic/real gap, nor uses such an analysis to inform how synthetic data is exploited for training.
\section{Experimental Setup}

\subsection{Models}

\begin{figure}[t]
  \centering
  \subfloat[]{\includegraphics[width=0.69\columnwidth]{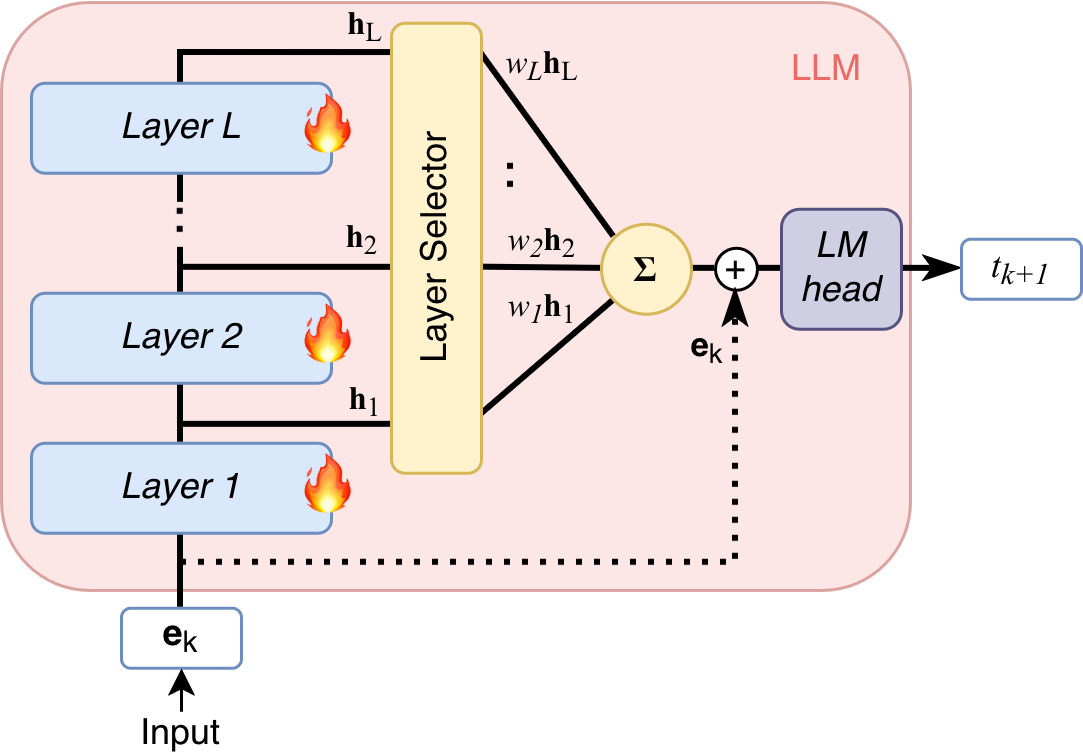}}\hfill
  \subfloat[]{\includegraphics[width=0.59\columnwidth]{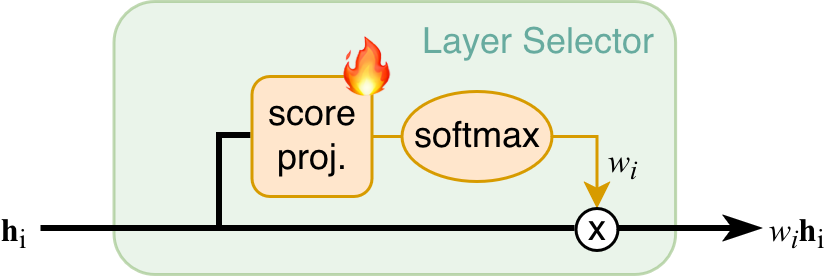}}
  \caption{Layer-wise Weighted Pooling inside of Llama architecture. All LLM hidden states (\texttt{L}) are weighted by a trainable parameter in order to select how each layer is kept before, optionally, the addition to the residual stream from the speech. Once done, it's passing through RMS~Norm and \texttt{lm\_head} to output the textual tokens of the transcripts.}
  \label{fig:lwp_arch}
\end{figure}

We build our systems on top of the SLAM-ASR framework~\cite{ma2024slam}, which couples a frozen WavLM-Large~\cite{chen2022wavlm} speech encoder to a Llama-3.2-3B-Instruct~\cite{dubey2024llama3} backbone through a single-hidden-layer projector that downsamples the audio representations by a factor of $5$. Domain adaptation uses LoRA~\cite{hu2022lora} adapters ($r=16$, $\alpha=32$, dropout 0.05) on the \texttt{q\_proj} and \texttt{v\_proj} modules of the LLM~(Figure \ref{fig:lwp_arch}), with all other parameters frozen. During the layer-wise analysis, we are probing the  architecture at its base checkpoint (projector trained, LLM frozen), before any domain fine-tuning. Representations are read directly from the 28 Llama layers, so the analysis reflects generic acoustic encoding rather than domain-specific adaptation.

\subsection{Datasets}


\noindent \textbf{Corpus.} \quad All experiments use DefinedAI~\cite{kumar2025distilling, carofilis2026textonly}, a corpus of manually transcribed English customer--agent telephone calls. The base model checkpoint is pre-trained from a mixed set of $\simeq$ 38 hours, giving a starting-point WER of 10.90\% on the held-out banking test set which is composed of 3,164 utterances (6.55 hours). All downstream fine-tuning and the layer-wise analysis use exclusively the banking training partition (26,457 utterances, 54.43 hours of real speech). Synthetic counterparts are generated with Qwen3-TTS~\cite{qwen3tts2025} for the same utterances, totalling $\simeq$ 51 hours with RIRs. The 100\%-real baseline reaches 8.68\% WER.
 
\noindent \textbf{Synthetic speech generation.} \quad
Synthetic utterances are produced with the Qwen3-TTS VoiceDesign variant, which is conditioned on a natural language voice prompt, rather than reference speaker audio. We selected Qwen3-TTS after preliminary comparative listening among the authors against several open-source alternatives \cite{du2024cosyvoice, casanova2024xtts, lacombe2024parlertts, zhou2025indextts2, omnivoice2025, chatterbox2025}, it gave the highest perceived naturalness, fewest artefacts, natural prosody and crucially, the most controllable voice characteristics, letting us match synthetic voices to real data persona metadata (gender, race) without any reference recordings. A per-role prompt is built from persona attributes and fixed style tokens (\emph{``clear articulation, naturally''}), yielding a distinct synthetic voice per speaker role in every dialog.

\noindent \textbf{Room Impulse Response Augmentation.} \quad
Convolving clean speech with measured room impulse responses (RIRs) is an established robustness augmentation~\cite{ko2017rir}. For synthetic speech specifically, RIR convolution injects the room reverberation and channel variability absent from pristine TTS recordings, masking their clean condition signature and narrowing the synthetic/real gap. We use the BUT Speech@FIT Reverb Database~\cite{szoke2019but} for this purpose.

\begin{figure}[H]
    \centering
    \includegraphics[width=0.8\columnwidth]{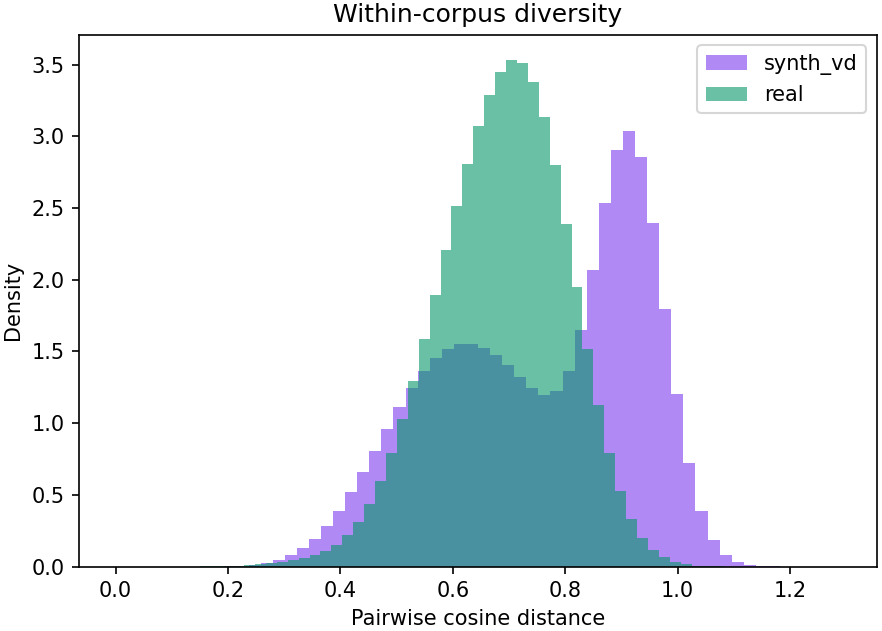}
    \caption{Within-corpus speaker diversity (pairwise cosine distance between \texttt{pyannote/embedding}~\cite{bredin2023pyannote} vectors; higher = more spread).}
    \label{fig:speaker_diversity}
\end{figure}

\begin{figure*}[t]
    \centering
    \includegraphics[width=\textwidth]{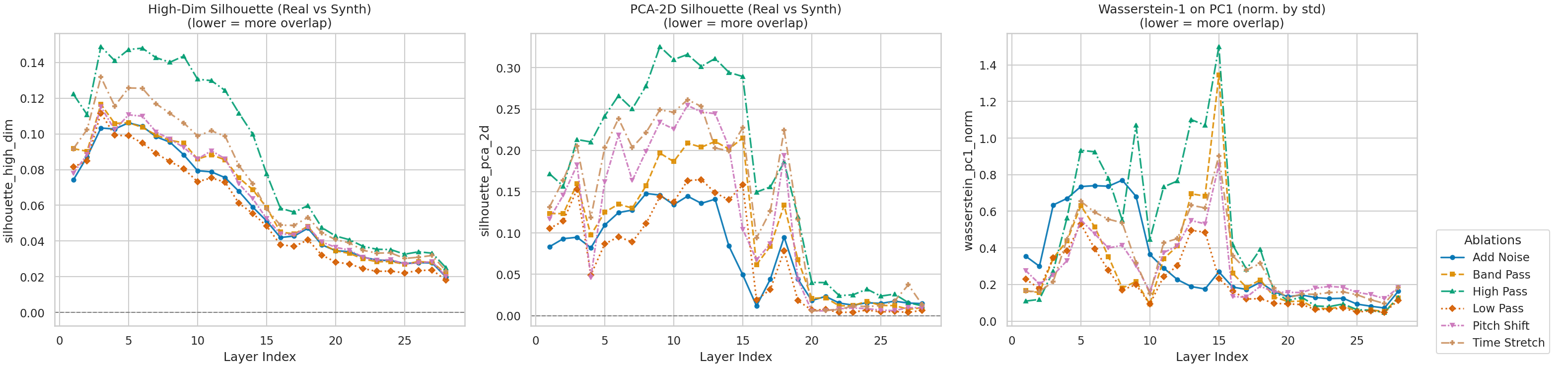}
    \caption{Layer-wise overlap metrics across all ablation conditions for 28 Llama layers. Lower values indicate greater real/synthetic overlap.}
    \label{fig:overlap}
\end{figure*}

\noindent \textbf{Speaker diversity.} \quad
Despite using no real speaker recordings, VoiceDesign yields a synthetic corpus with \emph{higher} within-corpus acoustic diversity than the real data. Figure~\ref{fig:speaker_diversity} shows pairwise cosine distances between \texttt{pyannote/embedding}~\cite{bredin2023pyannote} speaker vectors over all training utterances, the synthetic corpus (purple) is shifted right of real (green), dominant mode $\approx$ 0.88 vs.\ 0.68, i.e.\ more spread across speaker space. It is also bimodal, a secondary mode ($\approx$0.6) overlaps real, so synthetic keeps a real-like core while adding more diverse vocal characteristics than the acoustically homogeneous real corpus, thanks to generating a distinct voice per speaker role in each dialogue.

\begin{table}[H]
\centering
\caption{Audio quality of corpora (higher is better). UTMOS = Predicted MOS; PESQ = Audio quality measurement between the raw synthetic and the real audios.}
\label{tab:audio_metrics}
\begin{tabular}{lccc}
\toprule
\textbf{Condition} & \textbf{UTMOS} $\uparrow$ & \textbf{PESQ} $\uparrow$ \\
\midrule
Real               & 2.08 & -- \\
Synth (raw)        & 4.36 & 1.12 \\
Synth + RIRs       & 1.34 & 1.26 \\
\bottomrule
\end{tabular}
\end{table}

\noindent \textbf{Audio quality.} \quad
Table~\ref{tab:audio_metrics} reports UTMOS~\cite{saeki2022utmos} and wideband PESQ for the three conditions. Raw synthetic scores far higher than real on UTMOS (4.36 vs.\ 2.08), confirming Qwen3-TTS produces perceptually cleaner output than real telephone recordings.



RIR convolution sharply degrades naturalness (UTMOS 1.34), with only a marginal PESQ change against the synthetic source (1.26 vs.\ 1.12). The ASR gains come not from better naturalness but from making synthetic speech acoustically messier, like real telephone recordings, bridging the domain gap, not the perceptual one.

\subsection{Audio Filters}

We study six signal-level perturbations applied to synthetic audio, summarised with their hyperparameters in Table~\ref{tab:filter_params}. For the downstream ASR experiments we additionally evaluate four combined conditions: High/Low Pass each paired with Pitch Shift or Time Stretch.

\begin{table}[H]
\centering
\caption{Audio filter hyperparameters.}
\label{tab:filter_params}
\begin{tabular}{ll}
\toprule
\textbf{Filter} & \textbf{Hyperparameters} \\
\midrule
Add Noise    & Gaussian white noise, SNR $= 10$\,dB \\
Band Pass    & $f_\text{low}=300$\,Hz, $f_\text{high}=3{,}400$\,Hz \\
High Pass    & $f_c=400$\,Hz \\
Low Pass     & $f_c=2{,}000$\,Hz \\
Pitch Shift  & $\Delta p = +2$ semitones \\
Time Stretch & rate $= 1.2\times$ (pitch preserved) \\
\bottomrule
\end{tabular}
\end{table}

We further apply real-world RIR augmentation from the BUT Speech@FIT Reverb Database~\cite{szoke2019but} to synthetic data, adding room acoustics as a stronger form of domain adaptation than the signal-level perturbations alone.

\subsection{Training Protocol}
Training proceeds in two stages. \texttt{Stage~1} (projector pre-training) trains only the speech encoder projector on the full mixed-domains partition (38.17 hours). The LLM backbone and WavLM encoder are kept frozen. \texttt{Stage~2} (domain adaptation) loads the \texttt{Stage~1} projector and fine-tunes LoRA adapters on the banking specific data for each experimental condition. All components except the LoRA weights remain frozen. In the layer-wise weighted pooling experiments, the layer selector is additionally trained.
Both stages share the following hyperparameters: AdamW, learning rate $1{\times}10^{-4}$, linear warmup over 1,000 steps followed by linear decay to $0$, batch size $10$, gradient accumulation $1$, BF16 precision, random seed set at 42. All experiments are individually run on a single NVIDIA H100 GPU until reaching epoch 5.

\section{Where is Real/Synthetic Discrimination Encoded?}
\label{sec:ablation}

Before evaluating downstream ASR performance, we probe the model's encoder to localise \emph{where} and \emph{how strongly} the model separates real from synthetic speech, in order to know which audio perturbations are most effective at disrupting this separation.

\subsection{Overlap Metrics}
\label{subsection:overlap_metrics}

We quantify synthetic/real gap at each layer with four metrics (Figure~\ref{fig:overlap}, key layers in Table~\ref{tab:layer_filter_quant}). Two are silhouette scores, measuring class separation in the full representation space and after PCA projection to 2D (lower values mean more overlap). Wasserstein-1 on PC1, normalised by the pooled PC1 standard deviation for scale-invariance across layers, measures distributional distance along the first principal component (lower = more overlap). The PCA-2D KDE overlap coefficient fits a kernel density estimate per class in the 2D PCA space and integrates $\min(p_\text{real},\, p_\text{synth})$ over it (higher values, indicate greater overlap and less separation).


\subsection{Layer-wise Overlap Findings}

Table~\ref{tab:layer_filter_quant} reports the PCA-2D KDE overlap coefficient per filter at three representative Llama layers.

\begin{table}[H]
\centering
\caption{PCA-2D KDE overlap per filter at layers 3/14/28 (higher = higher overlap; bold = best per layer).}
\label{tab:layer_filter_quant}
\begin{tabular}{lccc}
\toprule
\textbf{Filter} & \textbf{Layer 3} $\uparrow$ & \textbf{Layer 14} $\uparrow$ & \textbf{Layer 28} $\uparrow$ \\
\midrule
Original     & 0.2304         & 0.2304         & 0.7835 \\
\midrule
Add Noise    & \textbf{0.458} & \textbf{0.675} & 0.754 \\
Low Pass     & 0.395          & 0.465          & \textbf{0.822} \\
Pitch Shift  & 0.338          & 0.557          & 0.809 \\
Band Pass    & 0.350          & 0.367          & 0.786 \\
Time Stretch & 0.288          & 0.290          & 0.793 \\
High Pass    & 0.243          & 0.261          & 0.759 \\
\bottomrule
\end{tabular}
\end{table}

We highlight four observations:
(1) Discrimination decreases monotonically with depth. Silhouette scores converge near zero by layer~28 across all filters, and the overlap coefficients in Table~\ref{tab:layer_filter_quant} reach 0.75 to 0.82 there, confirming that the final LLM layers collapse the synthetic/real gap regardless of perturbation. The discriminative signal therefore concentrates in early-to-middle layers (0 to 14).
(2) High Pass is the least effective filter. Across all metrics and layers, High Pass maintains the highest silhouette scores, so spectral attenuation of low frequencies alone is insufficient to confuse the decoder.
(3) Time Stretch achieves the broadest early-layer disruption. It produces the lowest silhouette at layer $\approx 3$, most effectively increasing discrimination in early representations; Pitch Shift and Band Pass achieve similar disruption at middle layers (layer $\approx 14$).
(4) Frequency-band filters match prosodic ones. Band Pass and Low Pass reduce discrimination nearly as well as Pitch Shift and Time Stretch in the mid-layer range, suggesting that restricting frequency content targets the same cues as prosodic modification.
These observations guide filter selection for the downstream experiments: Time Stretch and Pitch Shift are prioritised individually, and their combination with High Pass tests whether spectral and prosodic perturbations are complementary at the ASR level even when High Pass alone fails to disrupt the representations.

\section{Downstream ASR Experiments}
\label{sec:downstream}

\begin{table}[H]
\centering
\caption{Substitution: \%WER on real test set. Fixed budget variations the real/synth split at a constant total budget (real + synth $\approx$ 100\%); Fixed synth keeps the full synthetic set (100\%) and varies the real fraction.}
\label{tab:rq1}
\resizebox{\columnwidth}{!}{%
\begin{tabular}{lcc}
\toprule
\textbf{Training Composition} & \textbf{Synth Raw} $\downarrow$ & \textbf{Synth +RIRs} $\downarrow$ \\
\midrule
Base model ($\sim$40H mixed-domain) & \multicolumn{2}{c}{10.90} \\
100\% real / 0\% synth              & \multicolumn{2}{c}{8.68} \\
\midrule
\multicolumn{3}{l}{\emph{Fixed budget (real + synth $\approx$ 100\%)}} \\
90\% real / 10\% synth  & 9.10     & 8.46 \\
80\% real / 20\% synth  & 8.92     & 8.85 \\
70\% real / 30\% synth  & 8.66     & \textbf{8.45} \\
60\% real / 40\% synth  & 9.12     & 8.84 \\
50\% real / 50\% synth  & 10.08     & 9.31 \\
40\% real / 60\% synth  & 10.22     & 9.01 \\
30\% real / 70\% synth  & 9.81     & 9.24 \\
20\% real / 80\% synth  & 10.51     & 9.25 \\
10\% real / 90\% synth  & 11.00     & 10.26 \\
\midrule
\multicolumn{3}{l}{\emph{Fixed synth (100\% synth + X\% real)}} \\
10\% real / 100\% synth   & 10.33 & 9.83 \\
25\% real / 100\% synth   & 9.74  & \textbf{9.20} \\
50\% real / 100\% synth   & 9.19  & \textbf{8.28} \\
75\% real / 100\% synth   & 8.88  & 8.54 \\
100\% real / 100\% synth  & 8.74  & \textbf{8.01} \\
\bottomrule
\end{tabular}
}
\end{table}

The layer-wise analysis in Section~\ref{sec:ablation} shows that synthetic/real discrimination concentrates in the early-to-middle LLM layers and that temporal and prosodic perturbations disrupt it most. We now test whether these insights yield downstream gains, alongside three practical questions: how much real data synthetic audio can replace (\S\ref{sec:substitution}), how much synthetic data helps on top of full real data (\S\ref{sec:augmentation}), and whether interpretability-guided filters add benefit (\S\ref{sec:filters}). We also explore layer-wise weighted pooling, an architectural modification that aims to extract more value from imperfect synthetic data (\S\ref{sec:lwp}).

\subsection{How Much Real Data Can Be Substituted by Synthetic?}

\label{sec:substitution}
A key practical question is how much annotated real speech synthetic audio can replace without degrading ASR. We study this from two angles. The \emph{fixed-budget} regime varies the real/synthetic split in 10\% increments at a constant total budget (real + synth $\approx$ 100\%), with BUT RIRs throughout; the complementary \emph{fixed-synth} regime keeps the full synthetic set (100\%) and progressively adds real data, reporting both raw and RIR-augmented synthetic audio. Table~\ref{tab:rq1} reports \%WER on the real held-out test set.

The relationship is non-monotonic. Only the 90/10 (8.46\%) and 70/30 (8.45\%) mixtures with RIRs improve over the all-real baseline (8.68\%), whereas the 80/20 split (8.85\%) and all mixtures exceeding 40\% synthetic data degrade performance, reaching 9.31\% at equal proportions. Even in the extreme 10\%-real / 90\%-synthetic regime, however, the resulting system (10.26\%) still outperforms the unadapted base model (10.90\%), indicating that RIR-augmented synthetic data remains beneficial under severe data scarcity. We conclude that a modest synthetic proportion of 10 to 30\% is optimal. It matches or exceeds the all-real baseline while reducing the real-data requirement, whereas substituting more than 30\% of real recordings incurs a measurable WER penalty.

\subsection{How Much Synthetic Data Can Augment Real Speech?}
\label{sec:augmentation}

Rather than replacing real recordings, one may ask how much synthetic data can be added on top of the full real corpus. Table~\ref{tab:rq2} reports WER as the synthetic proportion grows while keeping all real data fixed.

\begin{table}[H]
\centering
\caption{Augmentation: \%WER on real test set as synthetic data is added to 100\% real speech, raw synthetic vs.\ RIR-augmented synthetic.}
\label{tab:rq2}
\begin{tabular}{lcc}
\toprule
\textbf{Training Composition} & \textbf{Synth Raw} $\downarrow$ & \textbf{Synth +RIRs} $\downarrow$ \\
\midrule
100\% real / 0\% synth   & 8.68  & -- \\
100\% real / 10\% synth  & 8.73  & 8.43 \\
100\% real / 25\% synth  & 8.88  & 8.28 \\
100\% real / 50\% synth  & 8.62  & 8.54 \\
100\% real / 75\% synth  & \textbf{8.56}  & 8.23 \\
100\% real / 100\% synth & 8.74  & \textbf{8.01} \\
\bottomrule
\end{tabular}
\end{table}

All RIR-augmented conditions outperform the real-only baseline. Notably, even 10\% synthetic with RIRs (8.43\%) outperforms any quantity of raw synthetic data. Raw synthetic data shows diminishing returns beyond 50\%, while RIR-augmented data consistently improves with scale up to 100\% synthetic. This confirms that acoustic environment diversity, not sheer data volume, is the driving factor.

\subsection{Do Interpretability-Guided Filters Improve ASR?}
\label{sec:filters}

The layer-wise analysis flagged Time Stretch, Pitch Shift, and Band Pass as effective at reducing synthetic/real separation, while High Pass alone was not. Table~\ref{tab:rq5} tests whether this translates to ASR gains on a 50\%-real / 50\%-synthetic split, varying the filter applied to synthetic audio. We add Low Pass and its combinations, since Low Pass achieved the highest layer 28 overlap in the interpretability analysis (Table~\ref{tab:layer_filter_quant}).

Among individual filters, Pitch Shift gives the largest gain ($-0.23\%$ WER without RIRs, $-0.40\%$ with), matching the interpretability finding that pitch is a key early-layer cue. Time Stretch without RIRs slightly hurts ($+0.15\%$), suggesting temporal distortion introduces mismatches that outweigh its layer-level masking benefit. Crucially, layer 28 overlap does not predict ASR benefit: all conditions converge there regardless of their effect at earlier layers, so the interpretability signal lives in the early-to-middle layers, not the final one. Among combinations, Low Pass + Time Stretch gives the best no-RIR result (8.68\%), matching the all-real baseline without any acoustic augmentation, but benefits little from RIRs. Low-pass filtering and reverberation appear to target overlapping cues. High Pass + Pitch Shift is instead the best condition with RIRs (8.58\%), retaining high-frequency content is more compatible with reverberation than suppressing it, indicating some filters act on cues not captured by the silhouette analysis.

\begin{table}[H]
\centering
\caption{\%WER on real test set. Effect of signal-level perturbations on ASR using 50\% real / 50\% synthetic training split.}
\label{tab:rq5}
\begin{tabular}{lcc}
\toprule
\textbf{Filter} & \textbf{Synth Raw} $\downarrow$ & \textbf{Synth +RIRs} $\downarrow$ \\
\midrule
None (baseline)              & 9.38  & 9.06 \\
\midrule
Pitch Shift                  & 9.15  & 8.66 \\
Time Stretch                 & 9.53  & 8.79 \\
\midrule
High Pass + Pitch Shift      & 9.14  & \textbf{8.58} \\
High Pass + Time Stretch     & 9.72  & 8.88 \\
Low Pass + Pitch Shift       & 8.74  & 9.10 \\
Low Pass + Time Stretch      & \textbf{8.68}  & 8.82 \\
\bottomrule
\end{tabular}
\end{table}

\subsection{Does Layer-wise Weighted Pooling Help in Low-Resource Settings?}
\label{sec:lwp}

\begin{table*}[h]
\centering
\caption{\%WER on real test set. Layer-wise weighted pooling (LWP) in the substitution setting (100\% synthetic). ``$-$res.'' drops the speech-token acoustic residual.}
\label{tab:rq4_rq11}
\begin{tabular}{lccccc}
\toprule
\textbf{Real fraction} & \textbf{w/o LWP, Raw} $\downarrow$ & \textbf{w/o LWP + RIRs} $\downarrow$ & \textbf{LWP, Raw} $\downarrow$ & \textbf{LWP + RIRs} $\downarrow$ & \textbf{LWP + RIRs ($-$res.)} $\downarrow$ \\
\midrule
\texttt{10\%} &  10.33 & 9.83 & 10.24 & \textbf{9.69} & 9.69 \\
\texttt{25\%} &   9.74 & 9.20 &  9.26 & \textbf{8.70} & 8.70 \\
\texttt{50\%} &   9.19 & \textbf{8.28} &  9.63 & \textbf{8.28} & 8.29 \\
\texttt{75\%} &   8.88 & 8.54 &  8.88 & \textbf{8.29} & \textbf{8.29} \\
\texttt{100\%} &   8.74 & \textbf{8.01} &  8.86 & 8.23 & 8.22 \\
\bottomrule
\end{tabular}
\end{table*}

\noindent \textbf{Mechanism.} \quad
The standard SLAM-ASR decoder feeds only the final LLM transformer layer's hidden states to language modeling head. We replace this with a layer-wise weighted pooling (LWP) module that learns, per token and per utterance, a softmax-weighted combination of all LLM layers:
\begin{equation}
  z(t) = \sum_l w_l(t)\,h_l(t), \quad
  w_l(t) = \operatorname{softmax}_l\!\bigl(\mathbf{s}^\top h_l(t)\bigr)
\end{equation}
where $\mathbf{s}\in\mathbb{R}^D$ is a zero-initialised score vector (the only new trainable parameter). Zero initialisation is a neutral prior, the initial softmax is a uniform $1/L$ mixture over layers, close to a simple average, without biasing toward any depth. An additional speech residual re-injects the projected encoder output at speech-token positions after pooling, preserving a direct acoustic information stream to the LM head.

\noindent \textbf{Substitution results.} \quad
Table~\ref{tab:rq4_rq11} compares \%WER for the substitution scenario with and without LWP. Without RIRs, LWP helps only at low real data fractions. $\mathbf{s}$ learns useful layer preferences from as little as 13.6\,h (25\%, $-0.48\%$ WER) but lacks signal at 10\% (5.4\,h) and adds nothing once real data is abundant. With RIRs the LWP is far stronger, at 25\% real, LWP+RIRs reaches 8.70\%, matching nearly the 100\% real baseline (8.68\%), and every setting with $\geq$ 25\% of real audios beats it (best 8.23\% at 100\% + 100\%). We attribute this to RIRs narrowing the acoustic gap, letting $\mathbf{s}$ learn stable preferences that transfer to the real test domain.

\noindent \textbf{Residual ablation.} \quad
Ablating the speech-token residual has no meaningful impact at any real data fraction, with $\leq$0.01\% absolute as shown in  Table~\ref{tab:rq4_rq11}. The pooling mechanism itself captures the relevant acoustic information and the residual stream is redundant since WavLM features already enter via the projector before the LLM layers and are reachable into initial layers of the LLM.

\noindent \textbf{Augmentation results.} \quad
Table~\ref{tab:rq_lwp_aug} shows that LWP without RIRs mostly hurts. With 54\,h of real data, the decoder already has well-calibrated last-layer representations, so the pooling weights only add noise. Adding RIRs recovers small gains at 10 to 50\% synthetic (e.g.\ 8.73\%~$\to$~8.37\% at 10\%) but degrades slightly at $\geq$75\%, where the processed synthetic utterances dominate and disrupt score learning. LWP therefore helps most in low-resource, substitution-heavy settings, not in augmentation, where abundant real data already anchors the decoder.

\begin{table}[h]
\centering
\caption{\%WER on real test set. Layer-wise weighted pooling in the augmentation setting (100\% real + X\% synthetic)}
\label{tab:rq_lwp_aug}
\begin{tabular}{lccc}
\toprule
\textbf{Synth fraction} & \textbf{w/o LWP, Raw} $\downarrow$ & \textbf{LWP, Raw} $\downarrow$ & \textbf{LWP + RIRs} $\downarrow$ \\
\midrule
10\%  & 8.73 & 8.61 & \textbf{8.37} \\
25\%  & 8.88 & 9.00 & \textbf{8.49} \\
50\%  & 8.62 & 8.75 & \textbf{8.43} \\
75\%  & 8.56 & 8.75 & \textbf{8.26} \\
100\% & 8.74 & 8.86 & \textbf{8.22} \\
\bottomrule
\end{tabular}
\end{table}

\noindent \textbf{Layer weight analysis.} \quad
Table~\ref{tab:lwp_weights} reports the mean softmax weights learned for the 25\%-real / 100\%-synthetic condition, by token type, on both real and synthetic test sets. The weights concentrate strikingly on layer 28, speech tokens receive mean weight 0.977 (synthetic) and 0.930 (real), far above the uniform $1/28\approx0.036$, with entropy as low as 0.161 nats (vs.\ 3.332 uniform) confirming near deterministic selection. Text tokens are less peaky (0.652/0.643; 1.303 nats), spreading residual weight near uniformly over layers 1 to 27.

This is consistent with the interpretability findings but answers a different question. Section~\ref{sec:ablation} showed early-to-middle layers (0 to 14) are most discriminative between real and synthetic. The LWP weights show the final layer is most useful for decoding, by layer 28 the model has refined acoustic input into a linguistically rich form optimal for transcription, regardless of origin. The two are complementary, early layers encode the domain gap, the final layer encodes what matters for output. The near identical profiles on both test sets confirm LWP learns domain-agnostic preferences, explaining why it generalises once RIRs reduce the acoustic gap.

\begin{table}[H]
\centering
\caption{Learned LWP attention on layer~28 by token type, for the 25\% real / 100\% synthetic model (uniform baseline $1/28\approx0.036$, test set, lower entropy = more concentrated).}
\label{tab:lwp_weights}
\begin{tabular}{lcccc}
\toprule
 & \multicolumn{2}{c}{\textbf{Layer-28 weight}} & \multicolumn{2}{c}{\textbf{Entropy (nats)}} \\
\cmidrule(lr){2-3}\cmidrule(lr){4-5}
\textbf{Token type} & \textbf{Synth} & \textbf{Real} & \textbf{Synth} & \textbf{Real} \\
\midrule
Speech  & 0.977 & 0.930 & 0.161 & 0.403 \\
Text    & 0.652 & 0.643 & 1.303 & 1.299 \\
All     & 0.834 & 0.820 & 0.661 & 0.768 \\
\midrule
Uniform & \multicolumn{2}{c}{0.036} & \multicolumn{2}{c}{3.332} \\
\bottomrule
\end{tabular}
\end{table}

\section{Conclusion}

The interpretability and ASR analyses prove complementary rather than redundant, synthetic/real discrimination concentrates in early-to-middle layers (0 to 14), yet LWP selects the final layer for decoding, early layers encode the domain gap while the final layer encodes the semantic part which matters for the task. RIR convolution dominates throughout, not by improving audio quality (UTMOS and PESQ show it makes synthetic audio less pristine) but by reproducing the acoustic irregularities of real telephone recordings, and LWP transfers to real audio only once RIRs supply this stable acoustic distribution.

Concretely, combining interpretability-guided filters (Pitch Shift, High Pass + Pitch Shift) with RIR augmentation reaches 8.01\% WER on the test set, beating the all-real baseline by 7.72\% relative (0.67\% absolute), and adding layer-wise weighted pooling lets a substitution-oriented system match that baseline with only 25\% real data (13.6\,h) and surpass it at every higher fraction.

Since filter effectiveness is likely tied to the encoder, future work will scale these findings to larger synthetic mixtures, additional TTS systems, alternative encoders, and multilingual settings. A further direction is to move beyond supervised fine-tuning and leverage synthetic data through reinforcement learning, optimising the ASR policy directly against WER. Recent critic-free methods such as Group Relative Policy Optimization (GRPO) \cite{shao2024deepseekmath} could exploit the large pool of synthetic and RIR-augmented utterances to extract further signal from synthetic data. All the code is available on GitHub.\footnote{Repository withheld for blind review.}

\section*{Acknowledgements}
This work was supported by Idiap Research Institute and Uniphore collaboration project. Part of this work was also supported by EU Horizon 2020 project ELOQUENCE (grant number 101070558).

\section{Generative AI Use Disclosure}
Specific sections were refined using Large Language Models to ensure linguistic clarity. All experiments, results, and scientific claims are the authors' own.



\bibliographystyle{IEEEtran}
\bibliography{references}

\begin{thebibliography}{10}
\providecommand{\url}[1]{#1}
\csname url@samestyle\endcsname
\providecommand{\newblock}{\relax}
\providecommand{\bibinfo}[2]{#2}
\providecommand{\BIBentrySTDinterwordspacing}{\spaceskip=0pt\relax}
\providecommand{\BIBentryALTinterwordstretchfactor}{4}
\providecommand{\BIBentryALTinterwordspacing}{\spaceskip=\fontdimen2\font plus
\BIBentryALTinterwordstretchfactor\fontdimen3\font minus \fontdimen4\font\relax}
\providecommand{\BIBforeignlanguage}[2]{{%
\expandafter\ifx\csname l@#1\endcsname\relax
\typeout{** WARNING: IEEEtran.bst: No hyphenation pattern has been}%
\typeout{** loaded for the language `#1'. Using the pattern for}%
\typeout{** the default language instead.}%
\else
\language=\csname l@#1\endcsname
\fi
#2}}
\providecommand{\BIBdecl}{\relax}
\BIBdecl

\bibitem{eu2024aiact}
{European Parliament and Council of the European Union}, ``Regulation (eu) 2024/1689 of the european parliament and of the council of 13 june 2024 laying down harmonised rules on artificial intelligence (artificial intelligence act),'' Official Journal of the European Union, OJ L, 2024/1689, 12.7.2024, 2024, \url{http://data.europa.eu/eli/reg/2024/1689/oj}.

\bibitem{guo2026explicitacousticevidenceperception}
\BIBentryALTinterwordspacing
X.~Guo, Y.~Xie, H.~Cheng, J.~Zhou, J.~Liu, H.~Huang, L.~Ye, and Q.~Zhang, ``Towards explicit acoustic evidence perception in audio llms for speech deepfake detection,'' 2026. [Online]. Available: \url{https://arxiv.org/abs/2601.23066}
\BIBentrySTDinterwordspacing

\bibitem{tang2024salmonn}
\BIBentryALTinterwordspacing
C.~Tang, W.~Yu, G.~Sun, X.~Chen, T.~Tan, W.~Li, L.~Lu, Z.~MA, and C.~Zhang, ``{SALMONN}: Towards generic hearing abilities for large language models,'' in \emph{The Twelfth International Conference on Learning Representations}, 2024. [Online]. Available: \url{https://openreview.net/forum?id=14rn7HpKVk}
\BIBentrySTDinterwordspacing

\bibitem{chu2023qwenaudioadvancinguniversalaudio}
\BIBentryALTinterwordspacing
Y.~Chu, J.~Xu, X.~Zhou, Q.~Yang, S.~Zhang, Z.~Yan, C.~Zhou, and J.~Zhou, ``Qwen-audio: Advancing universal audio understanding via unified large-scale audio-language models,'' 2023. [Online]. Available: \url{https://arxiv.org/abs/2311.07919}
\BIBentrySTDinterwordspacing

\bibitem{rubenstein2023audiopalmlargelanguagemodel}
\BIBentryALTinterwordspacing
P.~K. Rubenstein, C.~Asawaroengchai, D.~D. Nguyen, A.~Bapna, Z.~Borsos, F.~de~Chaumont~Quitry, P.~Chen, D.~E. Badawy, W.~Han, E.~Kharitonov, H.~Muckenhirn, D.~Padfield, J.~Qin, D.~Rozenberg, T.~Sainath, J.~Schalkwyk, M.~Sharifi, M.~T. Ramanovich, M.~Tagliasacchi, A.~Tudor, M.~Velimirović, D.~Vincent, J.~Yu, Y.~Wang, V.~Zayats, N.~Zeghidour, Y.~Zhang, Z.~Zhang, L.~Zilka, and C.~Frank, ``Audiopalm: A large language model that can speak and listen,'' 2023. [Online]. Available: \url{https://arxiv.org/abs/2306.12925}
\BIBentrySTDinterwordspacing

\bibitem{ma2024slam}
\BIBentryALTinterwordspacing
Z.~Ma, G.~Yang, Y.~Yang, Z.~Gao, J.~Wang, Z.~Du, F.~Yu, Q.~Chen, S.~Zheng, S.~Zhang, and X.~Chen, ``Speech recognition meets large language model: benchmarking, models, and exploration,'' in \emph{Proceedings of the Thirty-Ninth AAAI Conference on Artificial Intelligence and Thirty-Seventh Conference on Innovative Applications of Artificial Intelligence and Fifteenth Symposium on Educational Advances in Artificial Intelligence}, ser. AAAI'25/IAAI'25/EAAI'25.\hskip 1em plus 0.5em minus 0.4em\relax AAAI Press, 2025. [Online]. Available: \url{https://doi.org/10.1609/aaai.v39i23.34666}
\BIBentrySTDinterwordspacing

\bibitem{chen2022wavlm}
S.~Chen, C.~Wang, Z.~Chen, Y.~Wu, S.~Liu, Z.~Chen, J.~Li, N.~Kanda, T.~Yoshioka, X.~Xiao, J.~Wu, L.~Zhou, S.~Ren, Y.~Qian, Y.~Qian, J.~Wu, M.~Zeng, X.~Yu, and F.~Wei, ``Wavlm: Large-scale self-supervised pre-training for full stack speech processing,'' \emph{IEEE Journal of Selected Topics in Signal Processing}, vol.~16, no.~6, pp. 1505--1518, 2022.

\bibitem{radford2023whisper}
A.~Radford, J.~W. Kim, T.~Xu, G.~Brockman, C.~McLeavey, and I.~Sutskever, ``Robust speech recognition via large-scale weak supervision,'' in \emph{Proceedings of the 40th International Conference on Machine Learning}, ser. ICML'23.\hskip 1em plus 0.5em minus 0.4em\relax JMLR.org, 2023.

\bibitem{lu2026auditoryknowledge}
\BIBentryALTinterwordspacing
K.-H. Lu, S.-W. Fu, C.-H.~H. Yang, Z.~Chen, S.-F. Huang, C.-K. Yang, Y.-C. Lin, C.-Y. Hsiao, W.~Ren, E.-P. Hu, Y.-H. Huang, A.-Y. Cheng, C.-H. Chiang, Y.~Tsao, Y.-C.~F. Wang, and H.~yi~Lee, ``How auditory knowledge in llm backbones shapes audio language models: A holistic evaluation,'' 2026. [Online]. Available: \url{https://arxiv.org/abs/2603.19195}
\BIBentrySTDinterwordspacing

\bibitem{rossenbach2020ttsaugmentation}
N.~Rossenbach, A.~Zeyer, R.~Schlüter, and H.~Ney, ``Generating synthetic audio data for attention-based speech recognition systems,'' in \emph{ICASSP 2020 - 2020 IEEE International Conference on Acoustics, Speech and Signal Processing (ICASSP)}, 2020, pp. 7069--7073.

\bibitem{rosenberg2019synth}
A.~Rosenberg, Y.~Zhang, B.~Ramabhadran, Y.~Jia, P.~Moreno, Y.~Wu, and Z.~Wu, ``Speech recognition with augmented synthesized speech,'' in \emph{2019 IEEE Automatic Speech Recognition and Understanding Workshop (ASRU)}, 2019, pp. 996--1002.

\bibitem{wang21r_interspeech}
C.~Wang, A.~Wu, J.~Pino, A.~Baevski, M.~Auli, and A.~Conneau, ``{Large-Scale Self- and Semi-Supervised Learning for Speech Translation},'' in \emph{{Interspeech 2021}}, 2021, pp. 2242--2246.

\bibitem{yang2024versatiletts}
\BIBentryALTinterwordspacing
G.~Yang, F.~Yu, Z.~Ma, Z.~Du, Z.~Gao, S.~Zhang, and X.~Chen, ``Enhancing low-resource asr through versatile tts: Bridging the data gap,'' 2024. [Online]. Available: \url{https://arxiv.org/abs/2410.16726}
\BIBentrySTDinterwordspacing

\bibitem{perrin2025optsynth}
\BIBentryALTinterwordspacing
Y.~Perrin and G.~Boulianne, ``Towards improved speech recognition through optimized synthetic data generation,'' 2025. [Online]. Available: \url{https://arxiv.org/abs/2508.21631}
\BIBentrySTDinterwordspacing

\bibitem{srinivasavaradhan2025ttsstate}
P.~{Srinivasa Varadhan}, S.~Thomas, S.~{Teja M S}, S.~Bhooshan, and M.~M. Khapra, ``{The State Of TTS: A Case Study with Human Fooling Rates},'' in \emph{{Interspeech 2025}}, 2025, pp. 2285--2289.

\bibitem{su2024taskarith}
\BIBentryALTinterwordspacing
H.~Su, H.~Farn, F.-Y. Sun, S.-T. Chen, and H.-y. Lee, ``Task arithmetic can mitigate synthetic-to-real gap in automatic speech recognition,'' in \emph{Proceedings of the 2024 Conference on Empirical Methods in Natural Language Processing}, Y.~Al-Onaizan, M.~Bansal, and Y.-N. Chen, Eds.\hskip 1em plus 0.5em minus 0.4em\relax Miami, Florida, USA: Association for Computational Linguistics, Nov. 2024, pp. 8905--8915. [Online]. Available: \url{https://aclanthology.org/2024.emnlp-main.503/}
\BIBentrySTDinterwordspacing

\bibitem{chou2025selfrefining}
\BIBentryALTinterwordspacing
C.-K. Chou, C.-J. Hsu, H.-L. Chung, L.-H. Tseng, H.-C. Cheng, Y.-K. Fu, K.~P. Huang, and H.-Y. Lee, ``A self-refining framework for enhancing asr using tts-synthesized data,'' 2025. [Online]. Available: \url{https://arxiv.org/abs/2506.11130}
\BIBentrySTDinterwordspacing

\bibitem{tan2022naturalspeech}
X.~Tan, J.~Chen, H.~Liu, J.~Cong, C.~Zhang, Y.~Liu, X.~Wang, Y.~Leng, Y.~Yi, L.~He, S.~Zhao, T.~Qin, F.~Soong, and T.-Y. Liu, ``Naturalspeech: End-to-end text-to-speech synthesis with human-level quality,'' \emph{IEEE Transactions on Pattern Analysis and Machine Intelligence}, vol.~46, no.~6, pp. 4234--4245, 2024.

\bibitem{mishra2026explainable}
\BIBentryALTinterwordspacing
J.~Mishra, M.~Chhibber, H.~jin Shim, and T.~H. Kinnunen, ``Towards explainable spoofed speech attribution and detection:a probabilistic approach for characterizing speech synthesizer components,'' 2025. [Online]. Available: \url{https://arxiv.org/abs/2502.04049}
\BIBentrySTDinterwordspacing

\bibitem{wang2020asvspoof}
\BIBentryALTinterwordspacing
X.~Wang, J.~Yamagishi, M.~Todisco, H.~Delgado, A.~Nautsch, N.~Evans, M.~Sahidullah, V.~Vestman, T.~Kinnunen, K.~A. Lee, L.~Juvela, P.~Alku, Y.-H. Peng, H.-T. Hwang, Y.~Tsao, H.-M. Wang, S.~L. Maguer, M.~Becker, F.~Henderson, R.~Clark, Y.~Zhang, Q.~Wang, Y.~Jia, K.~Onuma, K.~Mushika, T.~Kaneda, Y.~Jiang, L.-J. Liu, Y.-C. Wu, W.-C. Huang, T.~Toda, K.~Tanaka, H.~Kameoka, I.~Steiner, D.~Matrouf, J.-F. Bonastre, A.~Govender, S.~Ronanki, J.-X. Zhang, and Z.-H. Ling, ``Asvspoof 2019: A large-scale public database of synthesized, converted and replayed speech,'' \emph{Computer Speech \& Language}, vol.~64, p. 101114, 2020. [Online]. Available: \url{https://www.sciencedirect.com/science/article/pii/S0885230820300474}
\BIBentrySTDinterwordspacing

\bibitem{dvirniak2026robustspeechdeepfakedetection}
\BIBentryALTinterwordspacing
A.~Dvirniak, E.~Kushnir, D.~Tarasov, A.~Iudin, O.~Kiriukhin, M.~Pautov, D.~Korzh, and O.~Y. Rogov, ``Towards robust speech deepfake detection via human-inspired reasoning,'' 2026. [Online]. Available: \url{https://arxiv.org/abs/2603.10725}
\BIBentrySTDinterwordspacing

\bibitem{baroudi24_interspeech}
S.~Baroudi, T.~Pellegrini, and H.~Bredin, ``{Specializing Self-Supervised Speech Representations for Speaker Segmentation},'' in \emph{{Interspeech 2024}}, 2024, pp. 3769--3773.

\bibitem{zaiem23b_interspeech}
S.~Zaiem, Y.~Kemiche, T.~Parcollet, S.~Essid, and M.~Ravanelli, ``{Speech Self-Supervised Representation Benchmarking: Are We Doing it Right?}'' in \emph{{Interspeech 2023}}, 2023, pp. 2873--2877.

\bibitem{11434629}
S.~Baroudi, H.~Bredin, J.~Razik, and R.~Marxer, ``On the use of self-supervised representation learning for speaker diarization and separation,'' in \emph{2025 IEEE Automatic Speech Recognition and Understanding Workshop (ASRU)}, 2025, pp. 1--7.

\bibitem{pasad2021layerwise}
A.~Pasad, J.-C. Chou, and K.~Livescu, ``Layer-wise analysis of a self-supervised speech representation model,'' in \emph{2021 IEEE Automatic Speech Recognition and Understanding Workshop (ASRU)}, 2021, pp. 914--921.

\bibitem{yang2021superb}
S.~wen Yang, P.-H. Chi, Y.-S. Chuang, C.-I.~J. Lai, K.~Lakhotia, Y.~Y. Lin, A.~T. Liu, J.~Shi, X.~Chang, G.-T. Lin, T.-H. Huang, W.-C. Tseng, K.~tik Lee, D.-R. Liu, Z.~Huang, S.~Dong, S.-W. Li, S.~Watanabe, A.~Mohamed, and H.~yi~Lee, ``{SUPERB: Speech Processing Universal PERformance Benchmark},'' in \emph{{Interspeech 2021}}, 2021, pp. 1194--1198.

\bibitem{hsu2026anatomymodalitygapdissecting}
\BIBentryALTinterwordspacing
M.-H. Hsu, X.~Zhang, X.~Tian, J.~Zhang, and Z.~Wu, ``Anatomy of the modality gap: Dissecting the internal states of end-to-end speech llms,'' 2026. [Online]. Available: \url{https://arxiv.org/abs/2603.01502}
\BIBentrySTDinterwordspacing

\bibitem{dubey2024llama3}
\BIBentryALTinterwordspacing
A.~Grattafiori \emph{et~al.}, ``The llama 3 herd of models,'' 2024. [Online]. Available: \url{https://arxiv.org/abs/2407.21783}
\BIBentrySTDinterwordspacing

\bibitem{hu2022lora}
\BIBentryALTinterwordspacing
E.~J. Hu, yelong shen, P.~Wallis, Z.~Allen-Zhu, Y.~Li, S.~Wang, L.~Wang, and W.~Chen, ``Lo{RA}: Low-rank adaptation of large language models,'' in \emph{International Conference on Learning Representations}, 2022. [Online]. Available: \url{https://openreview.net/forum?id=nZeVKeeFYf9}
\BIBentrySTDinterwordspacing

\bibitem{kumar2025distilling}
\BIBentryALTinterwordspacing
S.~Kumar, E.~Villatoro-Tello, S.~Burdisso, K.~Hacioglu, T.~Bañeras-Roux, H.~Watawana, D.~Sanchez-Cortes, S.~Madikeri, P.~Motlicek, and A.~Stolcke, ``Distilling conversations: Abstract compression of conversational audio context for llm-based asr,'' 2026. [Online]. Available: \url{https://arxiv.org/abs/2603.26246}
\BIBentrySTDinterwordspacing

\bibitem{carofilis2026textonly}
\BIBentryALTinterwordspacing
A.~Carofilis, S.~Burdisso, E.~Villatoro-Tello, S.~Kumar, K.~Hacioglu, S.~Madikeri, P.~Rangappa, M.~K. E, P.~Motlicek, S.~Venkatesan, and A.~Stolcke, ``Text-only adaptation in llm-based asr through text denoising,'' 2026. [Online]. Available: \url{https://arxiv.org/abs/2601.20900}
\BIBentrySTDinterwordspacing

\bibitem{qwen3tts2025}
\BIBentryALTinterwordspacing
H.~Hu, X.~Zhu, T.~He, D.~Guo, B.~Zhang, X.~Wang, Z.~Guo, Z.~Jiang, H.~Hao, Z.~Guo, X.~Zhang, P.~Zhang, B.~Yang, J.~Xu, J.~Zhou, and J.~Lin, ``Qwen3-tts technical report,'' 2026. [Online]. Available: \url{https://arxiv.org/abs/2601.15621}
\BIBentrySTDinterwordspacing

\bibitem{du2024cosyvoice}
\BIBentryALTinterwordspacing
Z.~Du, Q.~Chen, S.~Zhang, K.~Hu, H.~Lu, Y.~Yang, H.~Hu, S.~Zheng, Y.~Gu, Z.~Ma, Z.~Gao, and Z.~Yan, ``Cosyvoice: A scalable multilingual zero-shot text-to-speech synthesizer based on supervised semantic tokens,'' 2024. [Online]. Available: \url{https://arxiv.org/abs/2407.05407}
\BIBentrySTDinterwordspacing

\bibitem{casanova2024xtts}
E.~Casanova, K.~Davis, E.~Gölge, G.~Göknar, I.~Gulea, L.~Hart, A.~Aljafari, J.~Meyer, R.~Morais, S.~Olayemi, and J.~Weber, ``{XTTS: a Massively Multilingual Zero-Shot Text-to-Speech Model},'' in \emph{{Interspeech 2024}}, 2024, pp. 4978--4982.

\bibitem{lacombe2024parlertts}
Y.~Lacombe, V.~Srivastav, and S.~Gandhi, ``Parler-tts,'' \url{https://github.com/huggingface/parler-tts}, 2024.

\bibitem{zhou2025indextts2}
\BIBentryALTinterwordspacing
S.~Zhou, Y.~Zhou, Y.~He, X.~Zhou, J.~Wang, W.~Deng, and J.~Shu, ``Indextts2: A breakthrough in emotionally expressive and duration-controlled auto-regressive zero-shot text-to-speech,'' 2025. [Online]. Available: \url{https://arxiv.org/abs/2506.21619}
\BIBentrySTDinterwordspacing

\bibitem{omnivoice2025}
\BIBentryALTinterwordspacing
H.~Zhu, L.~Ye, W.~Kang, Z.~Yao, L.~Guo, F.~Kuang, Z.~Han, W.~Zhuang, L.~Lin, and D.~Povey, ``Omnivoice: Towards omnilingual zero-shot text-to-speech with diffusion language models,'' 2026. [Online]. Available: \url{https://arxiv.org/abs/2604.00688}
\BIBentrySTDinterwordspacing

\bibitem{chatterbox2025}
{Resemble AI}, ``{Chatterbox-TTS},'' \url{https://github.com/resemble-ai/chatterbox}, 2025, gitHub repository.

\bibitem{ko2017rir}
T.~Ko, V.~Peddinti, D.~Povey, M.~L. Seltzer, and S.~Khudanpur, ``A study on data augmentation of reverberant speech for robust speech recognition,'' in \emph{2017 IEEE International Conference on Acoustics, Speech and Signal Processing (ICASSP)}, 2017, pp. 5220--5224.

\bibitem{szoke2019but}
I.~Szöke, M.~Skácel, L.~Mošner, J.~Paliesek, and J.~Černocký, ``Building and evaluation of a real room impulse response dataset,'' \emph{IEEE Journal of Selected Topics in Signal Processing}, vol.~13, no.~4, pp. 863--876, 2019.

\bibitem{bredin2023pyannote}
H.~Bredin, ``{pyannote.audio 2.1 speaker diarization pipeline: principle, benchmark, and recipe},'' in \emph{{Interspeech 2023}}, 2023, pp. 1983--1987.

\bibitem{saeki2022utmos}
T.~Saeki, D.~Xin, W.~Nakata, T.~Koriyama, S.~Takamichi, and H.~Saruwatari, ``{UTMOS: UTokyo-SaruLab System for VoiceMOS Challenge 2022},'' in \emph{{Interspeech 2022}}, 2022, pp. 4521--4525.

\bibitem{shao2024deepseekmath}
\BIBentryALTinterwordspacing
Z.~Shao, P.~Wang, Q.~Zhu, R.~Xu, J.~Song, X.~Bi, H.~Zhang, M.~Zhang, Y.~K. Li, Y.~Wu, and D.~Guo, ``Deepseekmath: Pushing the limits of mathematical reasoning in open language models,'' 2024. [Online]. Available: \url{https://arxiv.org/abs/2402.03300}
\BIBentrySTDinterwordspacing

\end{thebibliography}

\end{document}